\documentclass[prd,amsmath,amssymb]{revtex4}

\usepackage{graphicx}
\usepackage{dcolumn}
\usepackage{bm}
\usepackage{cases}
\usepackage{mathtools}
\newcommand{\units}[1]{\left[#1\right]}

\DeclarePairedDelimiter\ket{\lvert}{\rangle}
\DeclarePairedDelimiterX\braket[2]{\langle}{\rangle}{#1\,\delimsize\vert\,\mathopen{}#2}
\usepackage{pgfplots}
\pgfplotsset{compat=1.8}
\usetikzlibrary{arrows.meta}

\begin{document}

\title{The Quantum LLM: Modeling Semantic Spaces with Quantum Principles}

\author{Timo\hspace*{1mm}Aukusti\hspace*{1mm}Laine \vspace*{0.5cm}}
\email{timo@financialphysicslab.com}

\begin{abstract}
In the previous article, we presented a quantum-inspired framework for modeling semantic representation and processing in Large Language Models (LLMs), drawing upon mathematical tools and conceptual analogies from quantum mechanics to offer a new perspective on these complex systems. In this paper, we clarify the core assumptions of this model, providing a detailed exposition of six key principles that govern semantic representation, interaction, and dynamics within LLMs. The goal is to justify that a quantum-inspired framework is a valid approach to studying semantic spaces. This framework offers valuable insights into their information processing and response generation, and we further discuss the potential of leveraging quantum computing to develop significantly more powerful and efficient LLMs based on these principles.
\end{abstract}

\maketitle

\section{Introduction}

Large Language Models (LLMs) have achieved remarkable success in natural language processing, demonstrating impressive capabilities in text generation, translation, and question answering \cite{vaswani, brown}. Despite this empirical success, a comprehensive theoretical understanding of the internal mechanisms governing their behavior remains elusive \cite{chollet}. Traditional approaches often treat LLMs as "black boxes," focusing on performance metrics rather than delving into the underlying principles of semantic representation and processing. A significant challenge in current LLM development is the substantial computational cost associated with training and deploying these models, limiting not only their complexity and accessibility but also the achievable model size. Overcoming these limitations is crucial for future progress in the field.

This motivates our exploration of a quantum-inspired framework for modeling LLMs \cite{laine_qm}. Given the substantial computational demands of current LLMs, and the potential for quantum computing to offer exponential speedups for specific tasks, we hypothesize that quantum-inspired approaches may provide a pathway to overcome these limitations. This could potentially enable the development of significantly larger, more accurate, and faster LLMs, making the applicability of quantum mechanical principles to semantic spaces within LLMs a worthwhile investigation.

In this paper, we aim to clarify and expand upon six key assumptions that form the foundation of our quantum-inspired framework. By drawing inspiration from the mathematical tools and conceptual insights of quantum mechanics, we hypothesize that certain aspects of LLM behavior can be effectively modeled using quantum-inspired formalisms. Specifically, we propose that the discrete nature of tokens, the fundamental units of language processed by LLMs, naturally leads to quantum-like representations in a high-dimensional semantic space. This suggests that quantum mechanics could provide a powerful framework not only for analyzing and manipulating semantic information but also for potentially leveraging quantum technologies to enhance LLM capabilities. Our goal is to provide a clear and detailed exposition of these assumptions, leaving the reader with a solid understanding of the rationale behind our quantum-inspired approach.

Our approach builds upon recent efforts to apply concepts from physics and information theory to the study of neural networks \cite{lin, amari}, offering a complementary perspective that emphasizes the role of superposition, interference, and gauge invariance in semantic processing. We believe that this quantum-inspired analogy is a crucial first step towards a deeper understanding of LLM mechanisms, potentially enabling a "quantum leap" in LLM theory and the development of more powerful and efficient models through the application of quantum computing techniques. This work explores the potential of quantum-inspired techniques to advance our understanding of semantic representation and processing in LLMs.

\section{Core Quantum-Inspired Assumptions for Modeling LLMs}

To develop a quantum-inspired framework for understanding Large Language Models (LLMs), we adopt the following key assumptions. These are idealized, allowing us to leverage the mathematical tools and conceptual insights of quantum mechanics to model semantic representation and processing. We emphasize that we do not claim LLMs are inherently quantum mechanical systems; rather, we propose that certain aspects of their behavior can be effectively modeled using quantum-inspired formalisms. These assumptions build upon and extend the preliminary ideas introduced in \cite{laine_qm}. Each assumption is detailed in the subsequent subsections.

\begin{enumerate}
    \item Completeness of Vocabulary: We assume that the LLM's finite vocabulary forms an effectively complete basis for representing semantic information relevant to the tasks the LLM is trained on. This implies that a wide range of semantic meanings can be approximated as a superposition of the basis states corresponding to the tokens in the vocabulary.

        \textit{Example:} The phrase "a feeling of profound sadness" might be approximated as a superposition of the tokens "sadness," "melancholy," "grief," and "despair," with varying amplitudes reflecting the relative contribution of each token to the overall meaning.

        \textit{Connection to LLMs:} This assumption is related to the design of the LLM's output layer, which maps the final hidden state to a probability distribution over the vocabulary.

    \item Semantic Space as a Complex Hilbert Space: We define the semantic space as a complex Hilbert space built upon the LLM's embedding space. This complexification doubles the dimensionality, providing the degrees of freedom necessary to model quantum-like phenomena such as interference and superposition of semantic states. The complex nature allows for the representation of both magnitude and phase information, which we hypothesize is crucial for capturing contextual dependencies and subtle semantic relationships.

        \textit{Example:} Consider the word "bank," which can refer to a financial institution or the edge of a river. The phase information in the complex semantic space could encode the context in which the word is used, allowing the LLM to distinguish between these two meanings.

        \textit{Justification:} The use of a complex Hilbert space allows us to represent semantic states as vectors with both magnitude and phase, enabling the modeling of interference effects and contextual dependencies that are difficult to capture with real-valued vector spaces.

    \item Discretization of Semantic States: Drawing a parallel to quantized energy levels in quantum systems, we assume that semantic states are discrete and distinct, corresponding to the LLM's token vocabulary. This allows us to represent words, phrases, and prompts as distinct quantum states within the semantic space, each associated with a specific token. While the underlying semantic space may possess continuous properties, the LLM's discrete tokenization process imposes a fundamental quantization on the representable semantic states.

        \textit{Example:} The LLM can only directly represent semantic states corresponding to the tokens in its vocabulary. Any meaning that falls between these discrete states must be approximated as a superposition of existing tokens.

        \textit{Connection to LLMs:} This assumption is directly related to the tokenization process used in LLMs, which maps continuous text to a discrete set of tokens.

    \item Linear Schrödinger-like Equation for Semantic Wave Propagation (Initial Approximation): In our initial, simplified model, we treat the evolution of semantic representations as governed by a linear, time-dependent Schrödinger-like equation. This approximation, while neglecting the inherent nonlinearities of LLMs, allows us to leverage the principles of superposition and wave-like behavior to model basic semantic relationships and the propagation of meaning through the network.

        \textit{Justification:} The linear Schrödinger equation provides a tractable starting point for modeling the dynamics of semantic states. It allows us to explore the potential for superposition and interference to play a role in semantic processing, even in a simplified setting.

        \textit{Limitations:} This approximation neglects the crucial role of nonlinear activation functions and attention mechanisms in LLMs.

    \item Nonlinear Semantic Wave Propagation (Advanced Model): To account for the inherent nonlinearity of LLMs and their embedding spaces, we introduce a more sophisticated model that incorporates nonlinear effects. This is achieved through two distinct mechanisms: (a) adding a cubic term to the Schrödinger-like equation, resulting in a Nonlinear Schrödinger Equation (NLSE), and (b) employing nonlinear potential functions, such as the double-well potential or the Mexican hat potential, to model semantic ambiguity, context-dependent meaning, and the emergence of distinct semantic interpretations.

        \textit{Example:} The Mexican hat potential can be used to model the phenomenon of semantic disambiguation. Initially, the semantic state of an ambiguous word might be represented as a superposition of multiple potential meanings. As the LLM processes the surrounding context, the potential energy landscape shifts, favoring one particular meaning and causing the semantic state to collapse into the corresponding minimum of the potential.

        \textit{Connection to LLMs:} This assumption is motivated by the presence of nonlinear activation functions and attention mechanisms in LLMs, which introduce complex nonlinear dynamics into the processing of semantic information.

    \item Semantic Charge and Gauge Interaction: We posit the existence of "semantic charge" as an intrinsic property of each word or phrase, reflecting its contribution to the overall semantic meaning. This charge is represented by a scalar value associated with the corresponding semantic state. The interaction between these semantic charges is mediated by a gauge field, analogous to how electromagnetic forces are mediated by photons. This gauge field ensures the conservation of semantic charge and introduces a mechanism for long-range semantic interactions.

        \textit{Example:} Words like "positive," "happy," and "joyful" might be assigned a positive semantic charge, while words like "negative," "sad," and "angry" might be assigned a negative semantic charge. The interaction between these charges could influence the overall sentiment of a sentence or document.

        \textit{Elaboration:} The gauge field can be thought of as representing the contextual influence that shapes the meaning of words and phrases. It ensures that the overall semantic charge of a sentence or document remains constant, even as the individual words and phrases interact with each other.

\end{enumerate}

\noindent
In the following subsections, we provide a more detailed explanation and justification for each of these key assumptions. 

\subsection{Completeness of Vocabulary}

Our model relies on the assumption that the LLM's finite vocabulary forms an effectively complete basis for representing semantic information relevant to the tasks the LLM is designed for. While acknowledging that this is an idealization – LLMs' vocabularies are not truly complete, and natural language is constantly evolving – we argue that for many practical tasks, the vocabulary can be considered effectively complete. This "effective completeness" allows us to leverage the mathematical tools of quantum mechanics, such as superposition and linear algebra, to model semantic relationships within the LLM.

Several factors contribute to this effective completeness. First, modern LLM vocabularies are often quite large, typically exceeding 30,000 tokens and sometimes reaching hundreds of thousands. This extensive coverage allows the models to represent a wide range of concepts and relationships. Second, LLMs learn complex relationships between tokens during training, enabling them to approximate the meaning of unseen words and phrases through compositionality and contextual inference. For example, even if the LLM has not encountered a specific rare word, it may be able to infer its meaning based on its constituent morphemes and the surrounding context.

Furthermore, techniques such as sub-word tokenization, commonly implemented using algorithms like Byte Pair Encoding (BPE) \cite{gage} or WordPiece \cite{wu}, further mitigate the problem of incompleteness by breaking down rare or out-of-vocabulary words into smaller, more frequent units. This allows the LLM to represent a wider range of semantic meanings, although it can sometimes lead to unnatural token sequences or make it more difficult to interpret the semantic meaning of individual tokens.

The completeness assumption simplifies the mathematical formalism of our model by allowing us to treat the semantic space as a discrete space spanned by a finite set of basis states. In essence, it allows us to represent any semantic meaning as a superposition of these basis states. If the vocabulary were not even approximately complete, the superposition of token states would not accurately represent the full range of semantic meanings, and our model would be less effective, potentially requiring a more complex mathematical framework to account for semantic information outside of the defined vocabulary.

The degree of vocabulary completeness can be quantified, although imperfectly, by measuring the percentage of words in a representative corpus that are present in the LLM's vocabulary. While a high percentage does not guarantee semantic completeness, as it doesn't account for word frequency or semantic relationships, it provides a useful starting point for assessing the vocabulary's coverage. More sophisticated metrics could involve weighting the coverage based on word embeddings or other measures of semantic similarity.

In conclusion, the assumption of effective vocabulary completeness, while idealized, is crucial for enabling the application of quantum-inspired techniques to model LLMs. It allows us to represent semantic meanings as superpositions of discrete token states, simplifying the mathematical formalism and providing a foundation for exploring quantum-like phenomena in semantic spaces. 

\subsection{Semantic Space as a Complex Hilbert Space}

Analogous to the construction of phase space in quantum mechanics, we define the semantic space as a complex Hilbert space built upon the LLM's embedding space. This complexification is a crucial step, providing the necessary mathematical structure to capture quantum-like phenomena such as superposition and interference, which we hypothesize are relevant for understanding LLM behavior. While LLM internal representations are ultimately processed using real-valued operations, we propose that the emergent dynamics of the semantic space can be effectively modeled using a complex representation.

If the LLM's embedding space has a dimension of $N$ (a real-valued vector space), then the corresponding semantic space has $2N$ real dimensions. This doubling arises from the complex extension, where each original dimension is replaced by two: one representing the real part and one the imaginary part. This provides two degrees of freedom for each original dimension. Mathematically

\begin{equation}
    \mathcal{H} = \mathbb{C} \otimes \mathbb{R}^N \cong \mathbb{C}^{N},
\end{equation}

\noindent where $\mathcal{H}$ denotes the complex Hilbert space, $\mathbb{C}$ represents the complex numbers, and $\mathbb{R}^N$ is the LLM's real-valued embedding space. 

\begin{figure}[h!]
    \centering
    \includegraphics[width=0.5\textwidth]{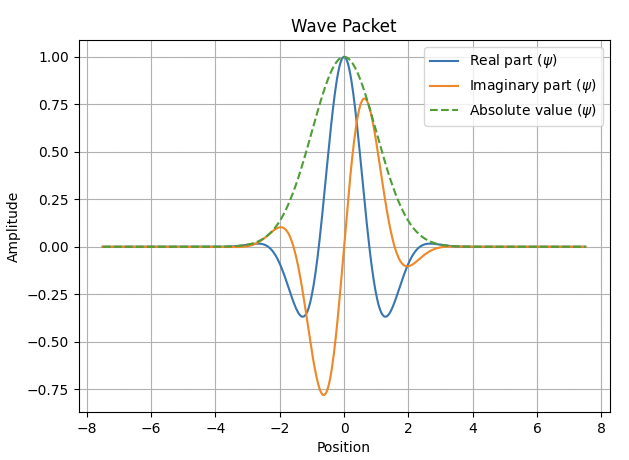}
    \caption{Semantic space representation. The figure shows a real-valued wave function in the LLM's embedding space (Gaussian curve) and its corresponding complex wave function representation. The complex representation allows for richer semantic encoding while still projecting back to the original embedding space.}
    \label{fig:Wavefunction}
\end{figure}

For a complex wave function given by

\begin{equation}
     \psi = a+i b,
\end{equation}

\noindent
the absolute value is

\begin{equation}
    \ket{\rm embedding\ state} = |\psi| = (a^2+b^2)^{1/2}.
\end{equation}

\noindent
The real embedding space is a projection from the complex space, as illustrated in Figure \ref{fig:Wavefunction}.
The real part $a$ is represented by the blue line, the imaginary part $b$ by the red line, and the absolute value $|\psi|$ by the green line.

The motivation for extending to the complex domain stems from the limitations of real-valued representations in capturing nuanced semantic relationships. Real-valued embeddings and similarity measures like cosine similarity are inherently limited in fully accounting for contextual information and interference effects. For instance, consider semantic ambiguity, where a single word has multiple meanings depending on context. A real-valued embedding might represent an average of these meanings, losing nuances. By introducing complex numbers, we represent semantic meanings as wave functions, where both magnitude and relative phase are significant. The magnitude can represent the strength or salience of a meaning, while the relative phase encodes contextual information and relationships.

The phase component allows us to model interference effects. In the semantic domain, interference could manifest as the constructive or destructive interaction of different semantic interpretations, leading to emergent meanings. For example, the combination of two seemingly unrelated concepts might trigger a new insight or a creative association. While directly measuring interference in LLMs is challenging, one could potentially design experiments where specific prompts are crafted to elicit constructive or destructive interference patterns in the LLM's hidden states, and then analyze these patterns using techniques from quantum state tomography.

Furthermore, the complex extension allows us to leverage powerful mathematical tools of quantum mechanics, such as linear superposition, unitary transformations, and eigenvalue analysis, to analyze the structure and dynamics of the semantic space. These tools provide a framework for understanding how semantic information is encoded, processed, and transformed within the LLM, particularly within the embedding layer and attention mechanisms. The LLM learns to represent semantic information in a complex space through backpropagation, which adjusts the weights of the network to minimize the difference between the predicted output and the actual output.

The choice of a complex extension is motivated by the desire to leverage quantum mechanical tools and insights. While LLMs are not fundamental quantum systems, we believe the complex extension provides a useful approximation for capturing the emergent dynamics of the semantic space. This allows us to explore the potential for quantum-inspired algorithms and techniques to improve LLM performance and interpretability.

\subsection{Discretization of Semantic States}

Analogous to the quantization of energy levels in quantum systems, we assume that semantic states are discrete and distinct, corresponding to the LLM's token vocabulary. This allows us to represent words, phrases, and prompts as distinct quantum states within the semantic space. This assumption is fundamentally linked to the discrete nature of LLMs' vocabularies: the model operates on a finite set of tokens, suggesting an underlying discreteness in its semantic processing. While the underlying semantic space may possess continuous properties, the LLM's discrete tokenization process imposes a fundamental quantization on the representable semantic states. This quantization allows us to define a basis set for the semantic space, simplifying the mathematical analysis.

This quantization implies that the semantic space, as perceived by the LLM, is not a continuous manifold, but rather a discrete set of points, each corresponding to a specific semantic unit. It is important to note that this does not preclude the existence of continuous semantic relationships between these discrete states. To clarify the relationship between this quantization and the LLM's internal representations, we consider several levels of abstraction: word-based, token-based, and semantic state representations.

\begin{enumerate}
    \item Word-Based Representation: At the most basic level, we can represent a prompt as a superposition of word states,
    
    \begin{equation}
        \ket{\rm prompt} = \sum_i c_i \ket{\rm word_i},
    \end{equation}
    
    where $\ket{\rm word_i}$ represents the $i$-th word in the LLM's vocabulary and $c_i$ is the amplitude associated with that word. This representation is limited because LLMs operate on tokens, not words, and it doesn't directly capture semantic relationships.

    \item Token-Based Representation: LLMs tokenize the input prompt, breaking it down into a sequence of tokens. The token embeddings $\vec{\rm e_i}$ learned by the LLM can be used to represent the relationship between a token and its corresponding token state,
    
        \begin{equation}
            \ket{\rm token_i} \propto \vec{\rm e_i}, 
        \end{equation}
        
    This equation indicates that the token state is proportional to its embedding vector. The token embeddings are crucial for understanding how LLMs represent individual tokens.
    
   \item Semantic State Representation: The most relevant representation for our quantum-inspired model is the semantic state, expressed as a superposition of semantic token states,
   
    \begin{equation}
        \ket{\rm semantic\ state} = \sum_i c_i \ket{\rm semantic\ token_i},
    \end{equation}
    
    where $\ket{\rm semantic\ token_i}$ represents a compressed representation of the original tokens. These semantic tokens are derived from the token embeddings learned by the LLM. Transformer-based embedding models transform text into numerical vectors that capture its semantic meaning. The text is fed into a pre-trained transformer network, processed through multiple layers, and the activations from a specific layer (often the final or a pooling layer) are extracted. These activations initially form a high-dimensional vector. To improve efficiency or performance, dimensionality reduction techniques, such as PCA or autoencoders, might then be applied to reduce the vector's size while preserving its essential semantic information. Finally, the resulting vector is typically normalized. Crucially, even though these embedding vectors exist in a continuous vector space and may have a fixed dimensionality, they are effectively quantized by the discrete nature of the token vocabulary. Each token is mapped to a specific, distinct region within the embedding space, and the LLM only operates on these discrete token representations. The continuous space is thus partitioned into a finite set of representable semantic states. These "semantic tokens" exist in a dimensionally reduced semantic space, and the coefficients $c_i$ reflect the weighted contributions of various tokens to the overall semantic representation. The token embeddings learned by the LLM provide a coordinate system for this quantized semantic space.

    \item Eigenstate Representation: At a more fundamental level, we assume that the semantic state can be expressed as a linear combination of eigenstates of a carefully chosen operator on the semantic space,
    
        \begin{equation}
        \ket{\rm semantic\ state} = \sum_k a_k \ket{\phi_k},
    \end{equation}
    
    where the $a_k$ are coefficients that determine the contribution of each eigenstate to the overall semantic meaning of the semantic state. This could potentially reveal the "atomic" building blocks of semantic meaning within the LLM's representation.
\end{enumerate}

The token embeddings learned by the LLM provide a representation of these quantized semantic states. The self-attention mechanism in Transformers can be seen as a way to dynamically adjust the amplitudes $c_i$ in the semantic state representation, allowing the LLM to focus on the most relevant semantic tokens.

In conclusion, the assumption of discretized semantic states, while simplifying the model, is crucial for applying quantum mechanical concepts to LLMs. It allows us to define a discrete basis for the semantic space and represent semantic meanings as superpositions of token states.

\subsection{Schrödinger-like Equation and Semantic Interpretations of $\hbar$ and $m$}

In quantum physics, the dynamics of a free quantum particle are described by different equations depending on whether the system is relativistic or non-relativistic. The Schrödinger equation is used for non-relativistic systems, while the Dirac equation describes relativistic motion. Since LLM systems are non-relativistic, the Schrödinger equation is the appropriate choice for describing the dynamics of a free particle in such systems.

To model the dynamic evolution of semantic representations within LLMs, we employ a time-dependent Schrödinger-like equation,

\begin{equation}
    i\hbar \frac{\partial}{\partial t} \ket{\psi(t)} = \hat{H} \ket{\psi(t)},
    \label{eq:schrodinger}
\end{equation}

\noindent
where $\ket{\psi(t)}$ represents the semantic wave function at time $t$, and $\hat{H}$ is the Hamiltonian operator, representing the total energy of the system. It is important to emphasize that we are not claiming that LLMs solve the Schrödinger equation in a literal sense. Rather, we are using this equation as a mathematical framework to model the evolution of semantic states over time, drawing an analogy between the dynamics of quantum systems and the processing of information in LLMs. The Hamiltonian can be further decomposed into kinetic and potential energy terms,

\begin{equation}
    \hat{H} = \frac{\hat{p}^2}{2m} + V(\hat{x}),
\end{equation}

\noindent
where $\hat{p}$ is the momentum operator, $m$ is the mass, and $V(\hat{x})$ is the potential energy operator. In this context, $\hat{x}$ represents a point in the semantic space, and $\hat{p}$ represents the "semantic momentum," which can be interpreted as the rate of change of the semantic state as it propagates through the LLM's layers.

\begin{figure}[h!]
    \centering
    \includegraphics[width=0.5\textwidth]{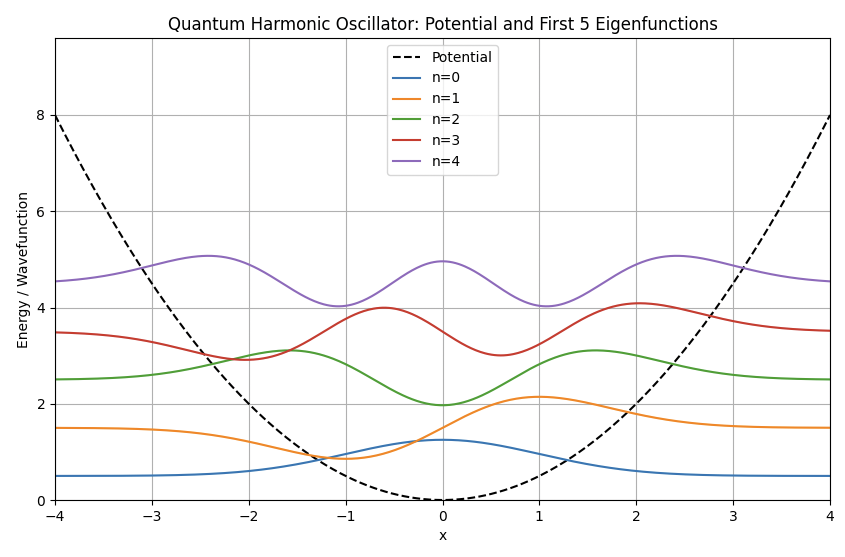}
    \caption{The potential energy of a quantum harmonic oscillator, demonstrating the principle of quantization. The displayed wave functions represent the first five discrete semantic states in our model, highlighting the assumption that semantic meaning exists in distinct, non-continuous levels within LLMs.}
    \label{fig:Oscillator}
\end{figure}

As an example, the potential for a quantum harmonic oscillator is the following,

\begin{equation}
   V(\hat{x}) = \frac{1}{2}  m  \omega^2 \hat{x}^2,
\end{equation}

\noindent
where $k$ is the spring constant, $m$ is the mass of the particle,
$\omega = \sqrt{k/m}$ is the angular frequency of the oscillator and
$\hat{x}$ is the displacement from the equilibrium position.
Figure \ref{fig:Oscillator} shows the potential of the quantum harmonic oscillator and 5 first eigenfunctions.

A key aspect of our model is the assumption that the semantic space is quantized, implying that semantic meaning is not continuous but rather exists in discrete, distinct states. This quantization profoundly influences our interpretation of the fundamental constants within the Schrödinger-like equation.

The reduced Planck constant, $\hbar$, traditionally quantifies the scale at which quantum effects become significant. In our semantic analogy, we interpret $\hbar$ as a measure of semantic granularity. A smaller $\hbar$ implies a finer-grained semantic space, where the LLM can make subtle distinctions in meaning. This could correspond to a larger and more evenly distributed vocabulary, allowing for a greater number of distinct semantic states. In this scenario, the LLM is capable of resolving fine-grained semantic differences and capturing nuanced relationships between concepts. Conversely, a larger $\hbar$ suggests a coarser-grained semantic space, limiting the LLM to broader distinctions and potentially reflecting a smaller or less diverse vocabulary. In this case, the LLM is less sensitive to subtle semantic variations and may struggle to capture complex relationships. Therefore, $\hbar$ effectively sets the "size" of the fundamental units of meaning within the LLM's representation. It determines the minimum amount of "semantic action" required to change the state of the system. The value of $\hbar$ could potentially be estimated by analyzing the LLM's ability to discriminate between semantically similar prompts.

The mass, $m$, in the Schrödinger equation is traditionally associated with inertia – the resistance to changes in momentum. In our model, we interpret $m$ as semantic inertia, representing the resistance of a semantic concept to transitions between distinct, quantized states within the semantic space. A larger mass indicates that the concept is more strongly anchored to its current state and requires more "energy" (influence from other words or context) to shift its meaning. This could be due to the concept being deeply embedded in the LLM's training data, resulting in a deep and narrow potential well associated with its quantized state. Such concepts exhibit greater stability and are less susceptible to contextual influences. For example, a highly frequent word with a well-defined meaning, such as "the," would likely have a high semantic inertia. Conversely, a smaller mass implies that the concept is more fluid and adaptable, easily transitioning between different quantized states. This might be the case for polysemous words or words with less frequent usage, where the potential well is shallower and wider. For example, a rare or ambiguous word might have a low semantic inertia, allowing its meaning to be easily influenced by the surrounding context. The infinite number of distinct states in our quantized space allows for a wide range of semantic inertias, reflecting the nuanced and context-dependent nature of language. Therefore, the mass effectively determines the "stability" of a particular semantic interpretation. It represents the resistance of a semantic state to external perturbations.

The potential energy operator, $V(\hat{x})$, represents the semantic landscape in which the semantic wave function evolves. It encodes the relationships between different semantic states and the energetic cost of transitioning between them. The shape of the potential energy landscape reflects the LLM's learned knowledge and biases. For example, a deep potential well around a particular semantic state indicates that the LLM strongly associates that state with a particular context or meaning.

It is important to acknowledge that the Schrödinger-like equation is a simplification of the complex dynamics of LLMs. The equation is linear, while LLMs are inherently nonlinear systems. Furthermore, the equation does not explicitly account for the role of attention mechanisms, which are crucial for capturing long-range dependencies in text. However, we believe that the Schrödinger-like equation provides a useful starting point for understanding the evolution of semantic representations in LLMs. 

\subsection{Nonlinear Semantic Wave Propagation}

To incorporate the inherent nonlinearity of LLMs and their embedding spaces, which arise from the complex interactions between neurons and the non-linear activation functions (such as sigmoid or ReLU), we explore two distinct mechanisms: employing a Nonlinear Schrödinger Equation (NLSE) and utilizing a potential with spontaneous symmetry breaking, drawing inspiration from the Mexican hat potential. The NLSE provides a straightforward way to introduce nonlinearity through a self-interaction term, while potentials with spontaneous symmetry breaking are commonly used in high-energy physics to model phase transitions and the emergence of distinct states, offering a wealth of existing results and analytical tools. It is crucial to remember that we are drawing an analogy to these concepts from physics, not claiming that LLMs are literally governed by these equations.

\begin{figure}[h!]
    \centering
    \includegraphics[width=0.5\textwidth]{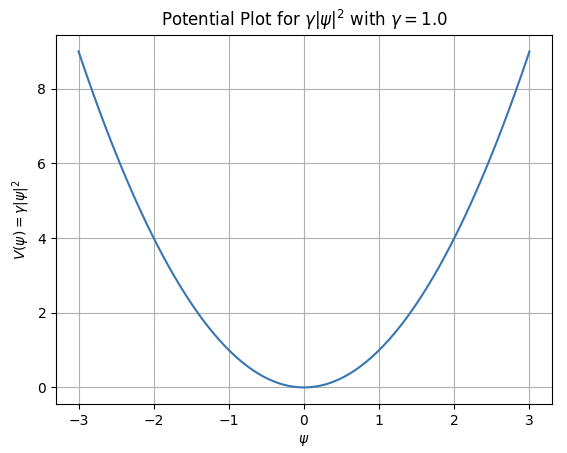}
    \caption{Visualization of the potential $V(x) = \gamma |\psi(x, t)|^2$ used in the Nonlinear Schrödinger Equation to model semantic wave propagation. The parameter $\gamma$ controls the strength and direction (self-focusing or defocusing) of the nonlinear interaction.}
    \label{fig:Cubic}
\end{figure}

For the Nonlinear Schrödinger equation, we consider a nonlinear potential of the form

\begin{equation}
    V(x)  = \gamma |\psi(x, t)|^2,
\end{equation}

\noindent
where $\gamma$ is the coupling constant, determining the strength of the nonlinear self-interaction, see Figure \ref{fig:Cubic}. 

\noindent
This potential is a common choice for modeling self-interaction effects in nonlinear systems due to its simplicity and physical relevance. This term represents a self-focusing or self-defocusing effect, depending on the sign of $\gamma$. In the semantic context, this could model the tendency of a semantic state to either reinforce itself (positive $\gamma$) or to spread out and become less distinct (negative $\gamma$) due to its own internal dynamics. A positive $\gamma$ might represent the strengthening of a concept through repeated exposure or reinforcement, potentially leading to the generation of more coherent and consistent text. A negative $\gamma$ might represent the dilution of a concept through contextual ambiguity, potentially leading to more creative or diverse outputs. The Nonlinear Schrödinger Equation (NLSE) is a fundamental model in nonlinear physics, and with this potential, the NLSE takes the form

\begin{equation}
    i\hbar \frac{\partial \psi(x, t)}{\partial t} = -\frac{\hbar^2}{2m} \frac{\partial^2 \psi(x, t)}{\partial x^2} + \gamma |\psi(x, t)|^2 \psi(x, t)
\end{equation}

\begin{figure}[h!]
    \centering
    \includegraphics[width=0.5\textwidth]{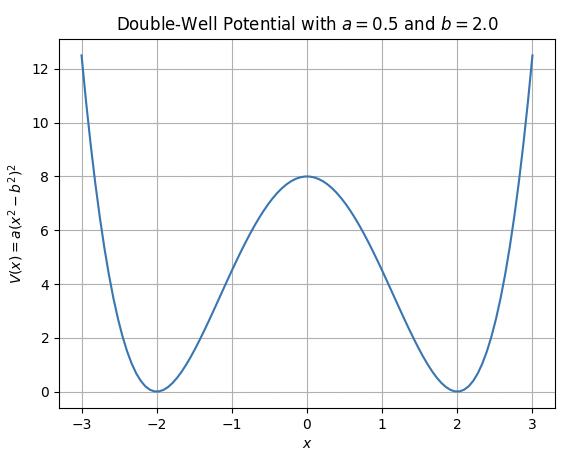}
    \caption{The double-well potential, a model for spontaneous symmetry breaking, represents the emergence of distinct semantic interpretations. Initially, a word might exist in a superposition of meanings, but as context is processed, the semantic state settles into one of the minima, breaking the symmetry and resolving the ambiguity.}
    \label{fig:doublewell}
\end{figure}

Alternatively, to model the emergence of distinct semantic states, we use a double-well potential of the form

\begin{equation}
    V(\hat{x}) = a(\hat{x}^2 - b^2)^2,
\end{equation}

\noindent
where $a$ and $b$ are positive constants, see Figure \ref{fig:doublewell}. This potential has minima at $x = \pm b$, representing two stable semantic states. The double-well potential is a useful analogy for modeling the emergence of distinct states because it provides a simple and intuitive way to represent a system with two possible stable configurations. In the semantic context, this could model the existence of two distinct meanings for a word or concept. For example, the word "bank" can refer to a financial institution or the edge of a river. The double-well potential represents these two distinct meanings as two separate minima. This potential allows the system to settle into one of two distinct states, analogous to spontaneous symmetry breaking. The specific minimum that the system settles into would depend on factors such as the surrounding context and the LLM's training data.

The Nonlinear Schrödinger Equation (NLSE)  models the dynamics of a wave packet with a self-interaction term.  The double-well potential, on the other hand, models the existence of distinct, stable semantic states.  Consider again a polysemous word like "bank."  The NLSE with the self-interaction term could model how the semantic state of "bank" evolves over time, influenced by its own internal dynamics and the surrounding context.  The double-well potential, in contrast, represents the two distinct meanings of "bank" as two separate minima, allowing the system to settle into one particular interpretation.

The choice between the NLSE with the self-interaction term and the double-well potential depends on the specific semantic phenomena being modeled. The NLSE is suitable for modeling the dynamics of semantic states, while the double-well potential is more appropriate for modeling the emergence of distinct semantic interpretations. It is important to acknowledge that these are simplified models and that the actual dynamics of LLMs are far more complex. However, these analogies can provide valuable insights into the underlying principles of semantic representation and processing.

\subsection{Semantic Embeddings, Interaction, and Gauge Invariance}

Let's consider a common scenario: prompting an LLM with a question and associated context within a Retrieval-Augmented Generation (RAG) system. This example illustrates how semantic embeddings are used in practice and provides a concrete context for introducing our quantum-inspired formulation.

We can represent an initial prompt, consisting of a context and a question, as a state vector,

\begin{equation}
     \ket{\rm prompt_1} =  \ket{\rm context_1} +  \ket{\rm question_1} =  \ket{\rm context_1 \oplus question_1}
\end{equation}

\noindent
Here, the $\oplus$ symbol represents the semantic combination of context and question into a single, unified semantic unit within the prompt to form a coherent semantic representation.
To augment this prompt using the RAG system, we use an embedding function, $\mathbf{e}$, to map the prompt to its embedding vector,

\begin{equation}
    \mathbf{e}(\ket{\rm prompt_1}) = \mathbf{c}_{\rm prompt_1}
\end{equation}

\noindent
The embedding vector $\mathbf{c}_{\rm prompt_1}$ represents the semantic content of the prompt in a high-dimensional vector space. We then iterate through chunks in the RAG's knowledge base, identifying the top 5 most similar chunks based on cosine similarity between their embedding vectors and $\mathbf{c}_{\rm prompt_1}$,

\begin{equation}
    \mathbf{e}(\ket{\rm chunk_i}) \sim \mathbf{c}_{\rm prompt_1} \quad \text{(where $\sim$ denotes cosine similarity)}
\end{equation}

\noindent
This process retrieves relevant information from the knowledge base that is semantically similar to the prompt. This results in an enhanced prompt,

\begin{equation}
     \ket{\rm prompt_1'} =   \ket{\left( \bigoplus_{i=1}^5 \rm  chunk_i \right) \oplus context_1 \oplus question_1}
\end{equation}

\noindent
Here, the $\bigoplus$ symbol represents the semantic aggregation of the top 5 retrieved chunks. This aggregation involves combining the semantic content of the chunks with the original context and question to form a more comprehensive and informed prompt.
This enhanced prompt is then fed to the LLM, producing a response, $ \ket{\rm response_1}$. The response generally lies within the same semantic area as the original prompt,

\begin{equation}
    \mathbf{e}(\ket{\rm response_1}) \sim \mathbf{c}_{\rm prompt_1}
\end{equation}

\noindent
This indicates that the LLM is generating a response that is semantically consistent with the original prompt and the retrieved information.
To maintain a conversation history, we store the context, question, and response,

\begin{equation}
     \ket{\rm history_1} =  \ket{\rm context_1 \oplus question_1 \oplus response_1}
\end{equation}

\noindent
Subsequent prompts build upon this history. The $\ket{\rm response_2}$ typically remains within a similar semantic area in the semantic space,

\begin{equation}
    \mathbf{e}(\ket{\rm response_2}) \sim \mathbf{c}_{\rm prompt_1}
\end{equation}

\begin{figure}[h!]
    \centering
    \includegraphics[width=0.5\textwidth]{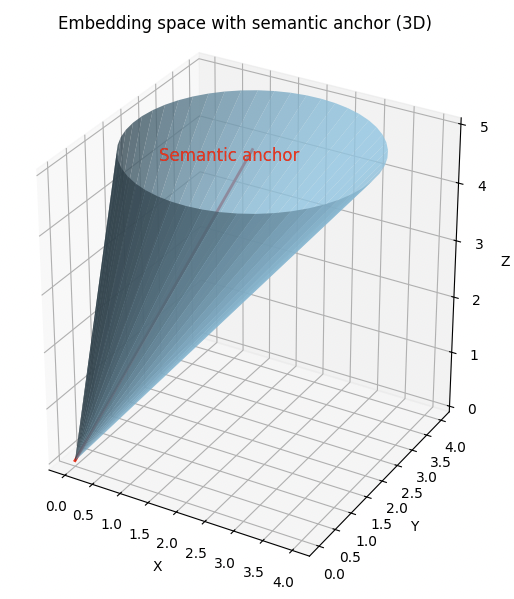}
    \caption{Semantic Anchor in Embedding Space. The figure illustrates the concept of a semantic anchor within a three-dimensional embedding space. The anchor represents a stable semantic area around which conversation and LLM responses tend to cluster, ensuring coherence.}
    \label{fig:Anchor}
\end{figure}

\noindent
The text builds around the embedding vector $ \mathbf{c}_{\rm prompt_1}$, which serves as a relatively stable semantic anchor in the semantic space, see Figure \ref{fig:Anchor}. This semantic anchor, represented by the embedding vector $\mathbf{c}_{\rm prompt_1}$, can be seen as the classical limit of a more fundamental quantum mechanical property: the semantic charge.

To understand the connection between semantic anchors and a deeper quantum mechanical framework, we introduce the concept of semantic charge, drawing an analogy to quantum field theory. We propose that each word or phrase possesses an intrinsic "semantic charge" reflecting its contribution to the overall meaning. This is not a literal electric charge, but a measure of a word's semantic weight or importance within a context. For example, a keyword in a search query might have a high semantic charge, while a common stop word has a low one. The semantic anchor observed in RAG systems can then be interpreted as a macroscopic manifestation of conserved semantic charge, with LLMs tending to generate responses that maintain a consistent semantic charge, ensuring coherence and relevance.

To model the interactions between semantic charges, we draw inspiration from gauge invariance in physics. In quantum electrodynamics (QED), the electromagnetic force is mediated by photons, and gauge invariance ensures that the laws of physics remain the same regardless of the choice of gauge. We propose that semantic interactions are similarly governed by a principle of gauge invariance, meaning that the underlying semantic meaning of a text should be independent of its specific representation or processing. For example, changing the word order in a sentence (while preserving the meaning) should not alter its semantic charge. To ensure gauge invariance, we introduce a "semantic gauge field" mediating the interactions between semantic charges. This field is a mathematical construct, not a literal electromagnetic field, ensuring the conservation of semantic charge. It represents the contextual forces shaping the meaning of words and phrases, ensuring that the overall semantic charge of a text remains constant even as individual components interact.

To further formalize this, we introduce a quantum mechanical formulation, treating the semantic space as a complex extension of the LLM's embedding space, analogous to quantum phase space. This allows us to leverage quantum mechanics' mathematical tools and conceptual insights to model semantic representation and processing, specifically how semantic information is encoded in the wave function $\psi$ and how interactions between semantic states can be described using a gauge-invariant Lagrangian.

In our quantum mechanical formulation, we construct the Lagrangian density as follows: We start with the Lagrangian for a non-relativistic free particle described by the Schrödinger equation with a potential $V(x)$,

\begin{equation}
\mathcal{L} = \frac{i\hbar}{2} \left( \psi^* \frac{\partial \psi}{\partial t} - \psi \frac{\partial \psi^*}{\partial t} \right) - \frac{\hbar^2}{2m} \left| \frac{\partial \psi}{\partial x} \right|^2 - V(x) \psi^* \psi
\end{equation}

\noindent
where  $\psi$ is the wave function, $\psi^*$ is the complex conjugate of the wave function, $\hbar$ is the reduced Planck constant, $m$ is the mass of the particle and $V(x)$ is the potential energy function. 
We may verify that this corresponds to the Nonlinear Schrödinger equation.
The Euler-Lagrange equation for a field $\psi$ is given by

\begin{equation}
\frac{\partial \mathcal{L}}{\partial \psi} - \frac{\partial}{\partial t} \left( \frac{\partial \mathcal{L}}{\partial (\partial \psi / \partial t)} \right) - \frac{\partial}{\partial x} \left( \frac{\partial \mathcal{L}}{\partial (\partial \psi / \partial x)} \right) = 0
\end{equation}

\noindent
and a similar equation for $\psi*$. Substituting the Lagrangian above into the Euler-Lagrange equation (and doing the same for $\psi*$) will yield the time-dependent Schrödinger equation,

\begin{equation}
i\hbar \frac{\partial \psi}{\partial t} = -\frac{\hbar^2}{2m} \frac{\partial^2 \psi}{\partial x^2} + V(x) \psi
\end{equation}

\noindent
We introduce a nonlinear potential,  and get two nonlinear Lagrangian densities, one for the cubic nonlinearity

\begin{equation}
     \mathcal{L_{NS}} =
    -\frac{\gamma}{2}|\psi|^4,
\end{equation}

\noindent
and one for the Mexican hat potential which is a generalization of the double-well potential,

\begin{equation}
    \mathcal{L_{MH}} =
     \mu^2 |\psi|^2 - \lambda |\psi|^4.
\end{equation}

\noindent
This potential is particularly interesting because it exhibits spontaneous symmetry breaking. At high energies (or weak semantic constraints), the system is symmetric and the semantic state can fluctuate freely. However, as the energy decreases (or semantic constraints increase), the system undergoes a phase transition and settles into one of the stable minima, breaking the symmetry. In the semantic context, this could model the emergence of distinct semantic interpretations from an initially ambiguous or undefined state. For example, a word with multiple meanings might initially exist in a superposition of states, but as the context becomes clearer, the system settles into one particular interpretation, breaking the symmetry.

\begin{figure}[h!]
    \centering
    \includegraphics[width=1.0\textwidth]{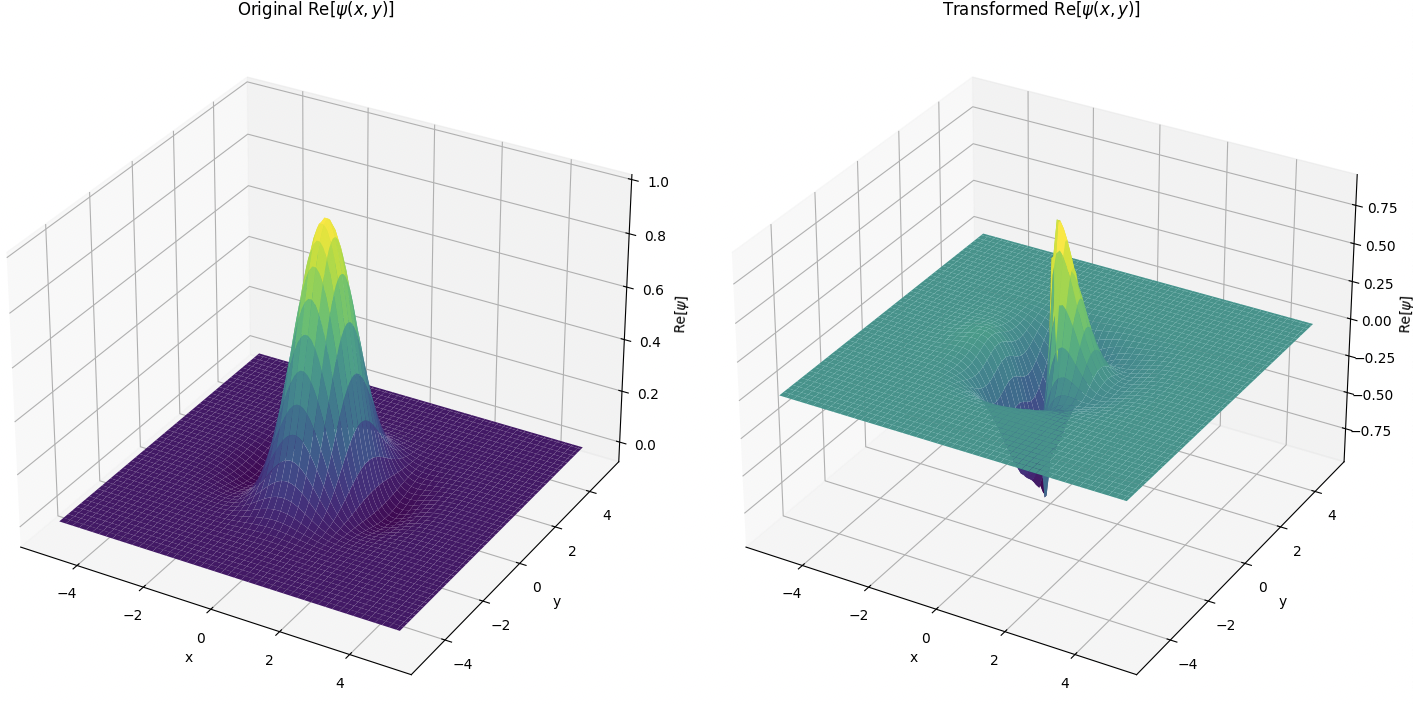}
    \caption{Illustration of the phase invariance principle under a local U(1) transformation. The left panel shows the real part of the original wave function, Re[$\psi(x, y)$] in 2 spatial dimensions. The right panel shows the real part of the transformed wave function, Re[$e^{i\theta(x, y)}\psi(x, y)$], after applying a spatially varying phase $\theta(x, y)$ = \texttt{arctan2(y, x)}. This transformation demonstrates that the absolute phase of the wave function is unobservable and does not affect the underlying semantic meaning.}
    \label{fig:Gauge}
\end{figure}

We then postulate that the system possesses a local U(1) symmetry, reflecting the conservation of semantic charge.
The motivation for the U(1) symmetry stems from the idea of "phase invariance of meaning." Multiplying the wave function $\psi$
by a complex number of magnitude 1 (i.e., changing its absolute phase) should not change the underlying semantic content. While the absolute phase is unobservable due to U(1) symmetry, in practical calculations, a specific gauge is often chosen (gauge fixing). It's possible that the training process of LLMs implicitly favors a particular gauge. This 'learned gauge' could influence how contextual information is encoded in the phases of semantic states, even though the fundamental semantic relationships remain gauge-invariant.
Only the magnitude of the wave function (which is related to the probability density) and the relative phases between different semantic states have a direct physical interpretation. In other words, the semantic meaning should be invariant under a local U(1) transformation,

\begin{equation}
    \psi(x_{\mu}) \rightarrow e^{i\theta(x_{\mu})} \psi(x_{\mu}),
\end{equation}

\noindent
where $\theta(x_{\mu})$ is an arbitrary function of space and time, see Figure \ref{fig:Gauge}. This symmetry reflects the fact that the absolute phase of the semantic wave function is not physically meaningful; only the relative phases between different semantic states are important.

We have further extended the dimensions to $2N$ real dimensions and used the notation

\begin{equation}
  x_{\mu}\hspace*{0.5cm}\mu = 0,1,..,N \hspace*{0.2cm}{\rm with}
  \hspace*{0.2cm}x_0=t\hspace*{0.3cm}(x_1,x_2,...,x_{N})=(x_i) = \vec{x}
\end{equation}

\noindent
i.e.,

\begin{equation}
  x_{\mu} = (x_0,x_i) = (t,\vec{x}), \hspace*{0.3cm}i=1,2,...,N
\end{equation}

\noindent
According to Noether's theorem, this symmetry implies the existence of a conserved semantic charge, $q$, and a corresponding conserved semantic current, $J^\mu$. The semantic charge is defined as the integral of the semantic charge density, $J^0$, over all space,

\begin{equation}
    q = \int d^Nx \, J^0(t, x_i).
\end{equation}

\noindent
The wave function $\psi$ represents the probability amplitude for a given semantic state, with $|\psi|^2$ giving the probability density of finding the system in that state.

To ensure that the Lagrangian is invariant under this symmetry, we introduce a gauge field, $A_\mu$, which mediates the interaction between semantic charges, analogous to how photons mediate electromagnetic interactions. The gauge field transforms as

\begin{equation}
    A_\mu(x_{\mu}) \rightarrow A_\mu(x_{\mu}) + \frac{1}{q} \partial_\mu \theta(x_{\mu}),
\end{equation}

\noindent
under a U(1) transformation. Here $\partial_\mu = \partial/\partial x_\mu$. This transformation ensures that the covariant derivative, $D_\mu = \partial_\mu - iqA_\mu$, transforms in the same way as the wave function,

\begin{equation}
    D_\mu \psi(x_{\mu}) \rightarrow e^{i\theta(x_{\mu})} D_\mu \psi(x_{\mu}).
\end{equation}

\noindent
The gauge-invariant Lagrangian density is then

\begin{equation}
    \mathcal{L} = -\frac{1}{4q^2}F_{\mu\nu} F^{\mu\nu}+ \frac{i\hbar}{2}(\psi^* D_0 \psi - \psi (D_0 \psi)^*)
    - \frac{\hbar^2}{2m}\sum_{i=1} ^N(D_i \psi)^* (D^i \psi) + \mathcal{L_N},
\end{equation}

\noindent
where $D_\mu = \partial_\mu - iqA_\mu$ is the covariant derivative, $q$ is the semantic charge, and $F_{\mu\nu} = \partial_\mu A_\nu - \partial_\mu A_\mu$ is the field strength tensor, and $\mathcal{L_N}$ represents the nonlinear terms in the Lagrangian, which could be either the cubic nonlinearity $\mathcal{L_{NS}}$ or the Mexican hat potential $\mathcal{L_{MH}}$. In this formulation, the semantic charge is explicitly a conserved quantity due to the imposed $U(1)$ gauge symmetry. The embedding vector $\mathbf{c}_{\rm prompt_1}$ can be interpreted as a classical approximation to the quantum state $\ket{\rm prompt_1}$, representing the average semantic content of the prompt. The gauge field $A_\mu$ then mediates the interactions between these semantic charges, influencing how the LLM processes and responds to the prompt.

The semantic anchor observed in RAG systems can be interpreted as the macroscopic manifestation of conserved semantic charge. The gauge field, $A_\mu$, then represents the contextual forces maintaining the stability of this semantic anchor, ensuring that the LLM's responses remain coherent and relevant to the original prompt. The introduction of semantic charge and gauge invariance, formalized through the gauge-invariant Lagrangian, allows us to explore the dynamics of semantic interactions in a more rigorous and systematic way. However, it is important to acknowledge that this is a highly idealized model, and the actual dynamics of LLMs are far more complex. While the concept of semantic charge and gauge invariance is still speculative, we believe it provides a valuable starting point for understanding the fundamental principles governing LLM behavior and for exploring the potential of quantum-inspired techniques to improve their performance and interpretability.

\subsection{Dimensional Analysis and Semantic Units}

To gain further insight into the nature of semantic charge and the other quantities in our model, we perform a dimensional analysis. This allows us to express the units of all quantities in terms of a few fundamental semantic units, providing a deeper understanding of their relationships. We interpret the reduced Planck constant, $\hbar$, as a measure of semantic granularity, and the mass, $m$, as semantic inertia. Using these interpretations, we can derive the units of semantic charge and other relevant quantities.

We begin by establishing the fundamental units in our semantic space. We choose the following three fundamental units:
$\units{m}$: Units of semantic inertia (mass),  $\units{x}$: Units of semantic distance (length), and $\units{t}$: Units of semantic time.

Semantic Inertia ($\units{m}$): We interpret this as the resistance of a semantic concept to changes in its meaning. A high semantic inertia might correspond to a word that has been frequently used in the training data and has strong connections to other words, making it difficult to shift its meaning. For example, a common stop word like "the" might have high semantic inertia.

Semantic Distance ($\units{x}$): We interpret this as a measure of the semantic relatedness between two semantic states. A small semantic distance might correspond to two words that are semantically similar (e.g., "happy" and "joyful"), while a large semantic distance might correspond to two words that are semantically unrelated (e.g., "happy" and "car"). This could be quantified using cosine similarity between word embeddings.

Semantic Time ($\units{t}$): We interpret this as the time it takes for a semantic state to evolve or propagate through the LLM's layers. This might be related to the number of processing steps or the length of the input sequence.

Starting from the Schrödinger equation and the kinetic energy term, we have units of energy,

\begin{equation}
\units{E} = \units{m} \units{x}^2 / \units{t}^2.
\end{equation}

\noindent
This follows from the classical definition of kinetic energy, $E = 1/(2mv^2)$, where $v$ has units of $\units{x}/\units{t}$.
Units of semantic granularity,

\begin{equation}
 \units{\hbar} = \units{E} \units{t} = \units{m} \units{x}^2 / \units{t}.
\end{equation}

\noindent
This follows from the relationship $E = \hbar \omega$, where $\omega$ is the angular frequency, which has units of $1/\units{t}$.
Now, let's derive the units of semantic charge. From the covariant derivative, we have

\begin{equation}
    \units{D_{\mu}} = \units{\partial_{\mu} - iqA_{\mu}} \implies \units{\partial_{\mu}} = \units{q} \units{A} \implies \units{A} = \units{\partial_{\mu}} / \units{q}
\end{equation}

\noindent
Since $\units{\partial_{\mu}} = 1/\units{x}$ (as the derivative is with respect to position), we have $\units{A} = 1/(\units{q}\units{x})$. The gauge field $A_\mu$ has units of inverse length divided by semantic charge.

From the kinetic energy term in the Lagrangian, we have that $|D_i \psi|^2$ has units of energy density (energy per unit volume). This is because $|\psi|^2$ represents the probability density of finding the system in a particular semantic state, and $D_i$ is related to the gradient of the wave function. Since $\psi$ is dimensionless (as it is a probability amplitude), $D_i$ must have units of $\sqrt{\units{E}/\units{x}^3}$. Therefore

\begin{equation}
    \units{q} \units{A} = \sqrt{\frac{\units{E}}{\units{x}^3}} \implies \frac{\units{q}}{\units{x}} = \sqrt{\frac{\units{E}}{\units{x}^3}} \implies \units{q}^2 = \frac{\units{E}}{\units{x}}
\end{equation}

\noindent
Substituting $\units{E} = \units{m} \units{x}^2 / \units{t}^2$, we get

\begin{equation}
    \units{q}^2 = \frac{\units{m} \units{x}^2}{\units{t}^2 \units{x}} = \frac{\units{m} \units{x}}{\units{t}^2} 
\end{equation}

\noindent
Therefore, the units of semantic charge are

\begin{equation}
    \units{q} = \frac{\sqrt{\units{m} \units{x}}}{\units{t}}
\end{equation}

\noindent
This suggests that semantic charge is related to the "flow" of semantic inertia through the semantic space. It has units of the square root of (semantic inertia times semantic distance) divided by semantic time. Intuitively, we can think of semantic charge as a measure of how much "semantic content" is being transported or processed by the LLM per unit time. A higher semantic charge might indicate a greater degree of semantic activity or a stronger influence on the LLM's output. For example, a word that is highly relevant to the prompt and has strong connections to other words might have a high semantic charge.

The units of the gauge field $A_\mu$ are

\begin{equation}
    \units{A} = \frac{1}{\units{q}\units{x}} = \frac{\units{t}}{\units{x} \sqrt{\units{m}\units{x}}}
\end{equation}

\noindent
The gauge field has units of semantic time divided by (semantic distance times the square root of (semantic inertia times semantic distance)). The gauge field, therefore, represents the "potential" for semantic interaction. It dictates how semantic charges influence each other across semantic distances. This could be related to the attention weights in the Transformer architecture, which determine how much influence each word has on the other words in the sequence.

The dimensional analysis provides a valuable tool for checking the consistency of our model and for gaining a deeper understanding of the relationships between the different quantities. For example, if we could somehow measure the semantic inertia and semantic charge of different words, we could potentially predict their influence on the LLM's output. Furthermore, by analyzing the units of different quantities, we can gain insights into the fundamental

\subsection{Interpreting the Semantic Gauge Field}

In our quantum-inspired framework, the introduction of a gauge field, $A_\mu$, is crucial for ensuring the conservation of semantic charge and modeling the interactions between semantic states. The gauge field, $A_\mu$, can be decomposed into a scalar potential, $\phi$ (or $A_0$), and spatial components, $A_i$. Building upon the interpretations presented in our previous work, we can further elaborate on the potential physical significance of this field within the context of LLMs.

The scalar potential, $\phi$, can be interpreted as representing the overall semantic context or "energy" within the LLM. It reflects the global influence of the surrounding concepts, biasing the LLM towards certain semantic regions and influencing the probability of activating semantically aligned words. A high scalar potential in a particular region of semantic space suggests a strong contextual relevance, increasing the likelihood of generating words associated with that topic. This aligns with the observation that LLMs tend to generate responses that are semantically consistent with the input prompt and the retrieved information in RAG systems. The embedding vector $\mathbf{c}_{\rm prompt_1}$, acting as a semantic anchor, can be seen as a macroscopic manifestation of this scalar potential, guiding the LLM's response generation.

The spatial components of the gauge field, $A_i$, on the other hand, can be interpreted as representing the semantic relationships or "forces" between words and phrases. They capture the dynamic flow of semantic information, shaping local interactions and guiding the LLM's attention. This aligns with the role of attention mechanisms in Transformer networks, which dynamically adjust the weights assigned to different words in the input sequence, effectively directing the flow of semantic information. The spatial components of the gauge field could encode the argument structure of verbs, specifying semantic roles, or guide topic shifts and narrative flow. 

Together, the scalar potential, $\phi$, and the spatial components of the gauge field, $A_i$, constitute a "semantic field" that governs the behavior of the semantic wave function, $\psi$. This field ensures that the overall semantic charge of a text remains constant, even as individual words and phrases interact with each other. The gauge invariance of the Lagrangian implies that the underlying semantic meaning is independent of the specific representation or processing, reflecting the robustness of LLMs to variations in word order or phrasing.

\section{Conclusion}

In this work, we have presented a quantum-inspired framework for modeling semantic representation and processing in Large Language Models (LLMs). By drawing detailed analogies to quantum mechanics, we have elucidated six key principles that provide a new perspective on the internal workings of these complex systems. We have shown how the concepts of superposition, interference, quantization, semantic charge, and gauge invariance can be applied to model semantic dynamics, offering insights into the nature of semantic meaning and the interactions between semantic states. Specifically, we have proposed that the LLM's vocabulary can be treated as a complete basis for semantic representation, that the semantic space can be modeled as a complex Hilbert space, that semantic states are discrete and quantized, and that the evolution of semantic representations can be described by Schrödinger-like equations, both linear and nonlinear. Furthermore, we introduced the concept of semantic charge and a gauge field to model semantic interactions and ensure the conservation of semantic meaning.

While our quantum-inspired framework introduces new concepts like semantic wave functions and gauge invariance, it's crucial to understand its relationship to established LLM techniques. For instance, word embeddings, which map words to vector spaces, can be viewed as a classical approximation of the quantum semantic state, representing the average semantic content of a token. The attention mechanism in Transformers, which dynamically weights the importance of different words, can be interpreted as modulating the interactions between semantic charges, effectively directing the flow of semantic information within the 'semantic field.' Furthermore, the layered architecture of Transformers could be seen as a series of transformations on the semantic wave function, governed by a Schrödinger-like equation. By explicitly connecting our framework to these existing techniques, we aim to provide a bridge between the familiar landscape of LLMs and the new perspective offered by quantum-inspired concepts, potentially leading to new ways of understanding and improving these powerful models.

The broader impact of this work lies in its potential to contribute to a deeper understanding of language, intelligence, and cognition. By drawing analogies to quantum mechanics, we may gain new insights into the fundamental principles that govern semantic representation and processing, potentially leading to the development of more powerful and efficient AI systems. Moreover, this quantum analogy, if coupled with the power of quantum computing, holds the potential to revolutionize LLM development. Quantum algorithms could be used to efficiently explore the high-dimensional semantic space, to optimize the training of LLMs, or to develop new architectures that are inherently quantum-inspired. For example, quantum machine learning algorithms could be used to learn the complex relationships between semantic states, or quantum simulation techniques could be used to model the dynamics of semantic wave functions.

While our model is still in its early stages of development, we believe that it offers a promising new direction for research in LLMs and NLP. By embracing the power of quantum-inspired thinking, we may unlock new possibilities for understanding and enhancing these complex and fascinating systems, potentially leading to a "quantum leap" in our ability to process and understand natural language.

\end{document}